\documentclass[10pt,twocolumn,letterpaper]{article}

\usepackage{cvpr}
\usepackage{times}
\usepackage{epsfig}
\usepackage{graphicx}
\usepackage{amsmath}
\usepackage{amssymb}
\usepackage{authblk}
\usepackage{subfig}
\usepackage{array}
\usepackage{cite}
\newcolumntype{P}[1]{>{\centering\arraybackslash}p{#1}}
\newcolumntype{M}[1]{>{\centering\arraybackslash}m{#1}}

\usepackage[pagebackref=true,breaklinks=true,letterpaper=true,colorlinks,bookmarks=false]{hyperref}

\cvprfinalcopy 

\def\cvprPaperID{248} 

\ifcvprfinal\fi
\begin{document}

\title{Differential Angular Imaging for Material Recognition}

\author[1]{
Jia Xue
}
\author[1]{
Hang Zhang
}
\author[1]{
Kristin Dana
}
\author[2]{
Ko Nishino
}
\affil[1]{Department of Electrical and Computer Engineering, Rutgers University, Piscataway, NJ 08854}
\affil[2]{Department of Computer Science, Drexel University,  Philadelphia, PA 19104}
\affil[ ]{ {\tt\small \{jia.xue,zhang.hang\}@rutgers.edu, kdana@ece.rutgers.edu,  kon@drexel.edu}}

\renewcommand\Authsep{  } 
\renewcommand\Authands{  }


\maketitle
\thispagestyle{empty}

\begin{abstract}
Material recognition for real-world outdoor surfaces has become increasingly important for computer vision to support its operation ``in the wild.''
Computational surface modeling that underlies material recognition has transitioned from reflectance modeling using in-lab controlled radiometric measurements to image-based representations based on internet-mined images of materials captured in the scene. We propose  a middle-ground approach that takes advantage of both rich radiometric cues and flexible image capture. 
We develop a framework for differential angular imaging, where small angular variations in image capture provide an enhanced appearance representation and significant recognition improvement. 
We build a large-scale material database, Ground Terrain in Outdoor Scenes (GTOS), geared towards real use for autonomous agents. This publicly available database\footnote{\url{http://ece.rutgers.edu/~kdana/gts/gtos.html}} consists of over 30,000 images covering 40 classes of outdoor ground terrain under varying weather and lighting conditions.
We develop a novel approach for material recognition called a Differential Angular Imaging Network (DAIN) to fully leverage this large dataset. With this network architecture, we extract characteristics of materials encoded in the angular and spatial gradients of their appearance. Our results show that DAIN achieves recognition performance that surpasses single view or coarsely quantized multiview images. These results demonstrate the effectiveness of differential angular imaging as a means for flexible, in-place material recognition. 
\end{abstract}

\vspace{-0.2in}
\section{Introduction}
\begin{figure}
\centering
\includegraphics[width= .9\linewidth]{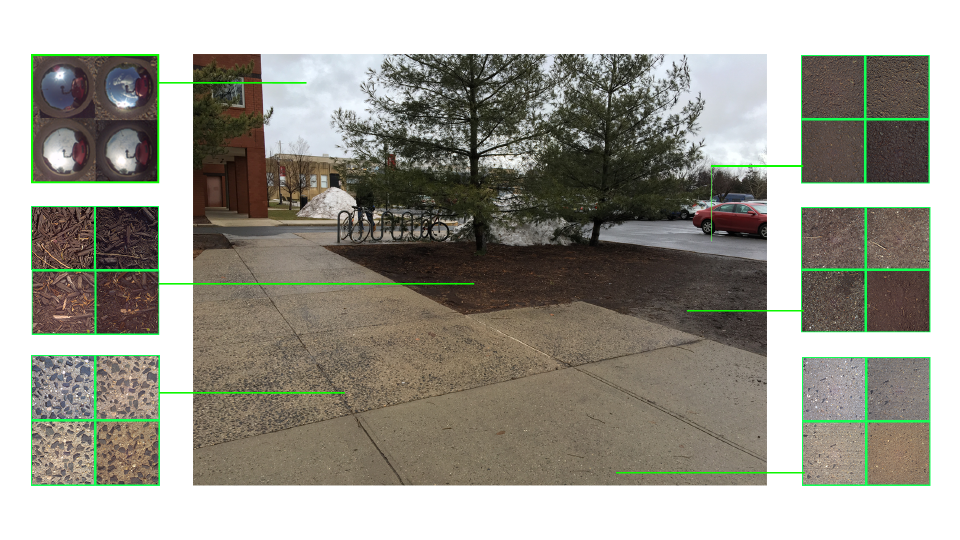}
\includegraphics[width= .9\linewidth]{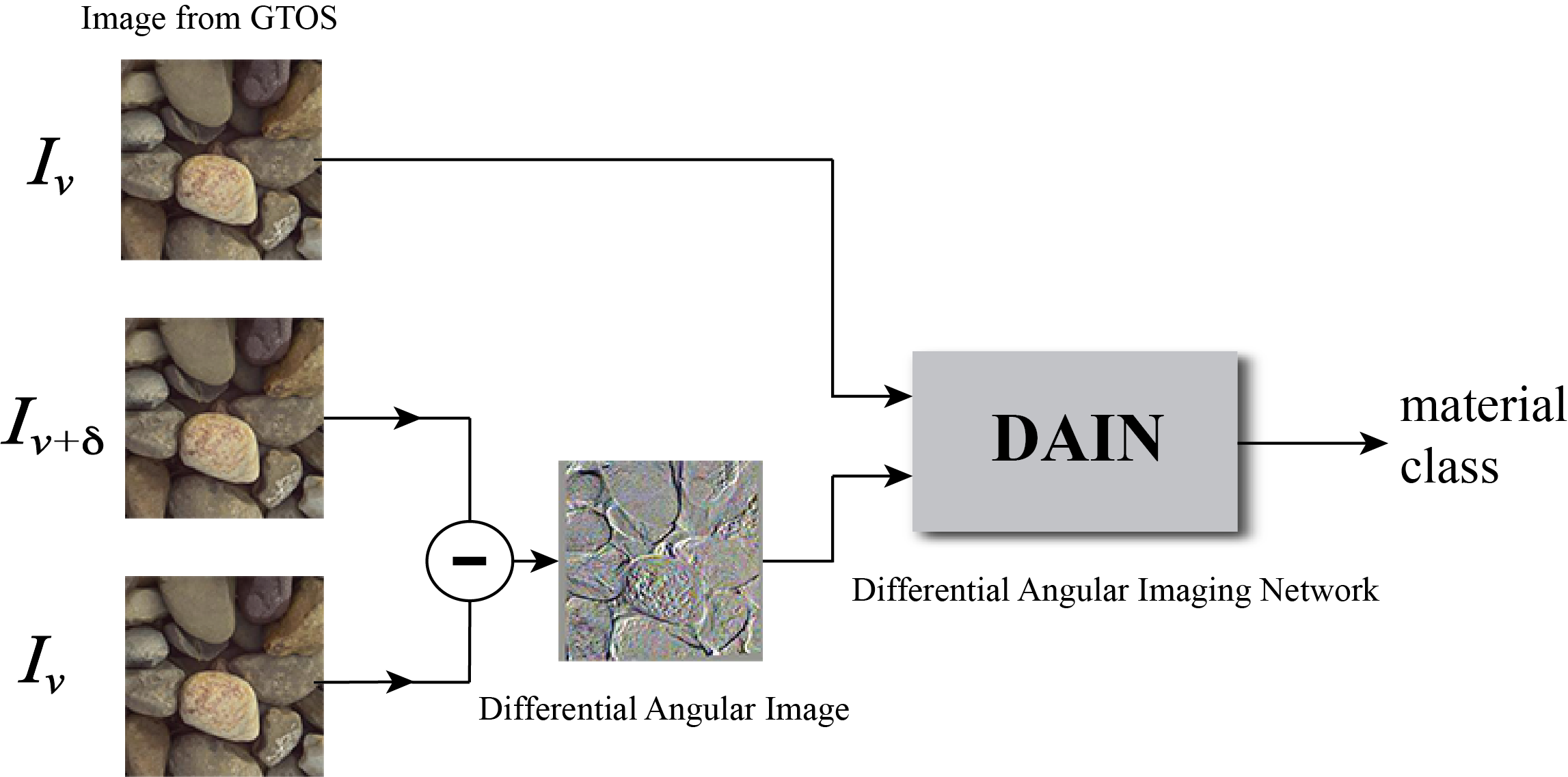}
\caption{(Top) Example from GTOS dataset comprising   outdoor measurements with multiple viewpoints, illumination conditions and angular differential imaging.  The example shows scene-surfaces imaged  at different illumination/weather conditions.  (Bottom)  Differential Angular Imaging Network (DAIN) for material recognition. }
\label{fig:figure1}
\vspace{-0.15in}
\end{figure}

 \begin{figure*}
\centering
\subfloat
{
\includegraphics[width=.13\linewidth]{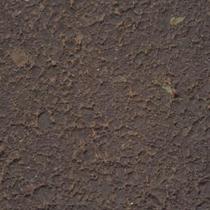}
}
\subfloat
{
\includegraphics[width=.13\linewidth]{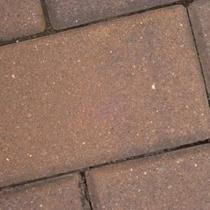}
}
\subfloat
{
\includegraphics[width=.13\linewidth]{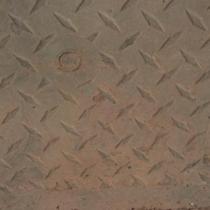}
}
\subfloat
{
\includegraphics[width=.13\linewidth]{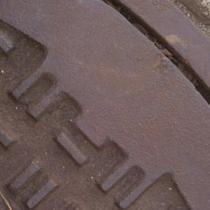}
}
\subfloat
{
\includegraphics[width=.13\linewidth]{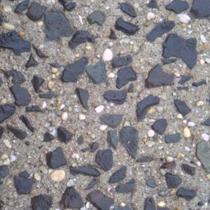}
}
\subfloat
{
\includegraphics[width=.13\linewidth]{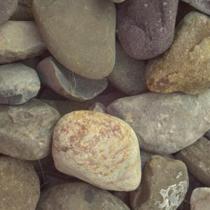}
}
\subfloat
{
\includegraphics[width=.13\linewidth]{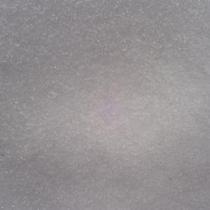}
}
\setcounter{subfigure}{0}
\subfloat[Asphalt]
{
\includegraphics[width=.13\linewidth]{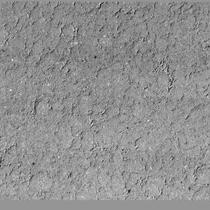}
}
\subfloat[Brick]
{
\includegraphics[width=.13\linewidth]{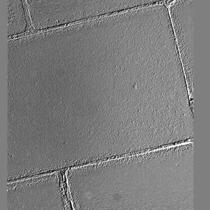}
}
\subfloat[Plastic cover]
{
\includegraphics[width=.13\linewidth]{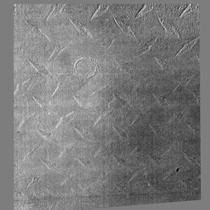}
}
\subfloat[Metal cover]
{
\includegraphics[width=.13\linewidth]{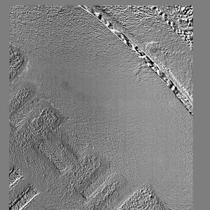}
}
\subfloat[Stone-cement]
{
\includegraphics[width=.13\linewidth]{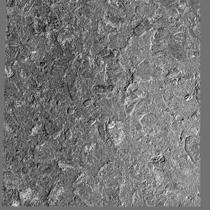}
}
\subfloat[Pebble]
{
\includegraphics[width=.13\linewidth]{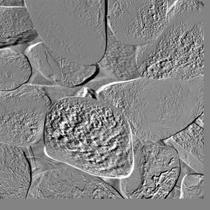}
}
\subfloat[Snow]
{
\includegraphics[width=.13\linewidth]{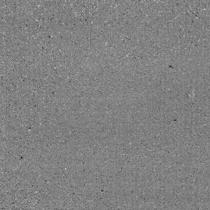}
}
\caption{Differential Angular Imaging. (Top) Examples of material surface images $I_v$. (Bottom) Corresponding  differential images $I_\delta = I_v - I_{v+\delta}$ in our GTOS dataset. These sparse images encode angular gradients of reflection and 3D relief texture.}
\label{fig:compare_img_patch}
\vspace{-0.2in}
\end{figure*}

Real world scenes consist of surfaces made of numerous materials, such as wood, marble, dirt, metal, ceramic and fabric, which contribute to the rich visual variation we find in images. 
Material recognition has become an active area of research in recent years
with the goal of providing detailed material information for applications such as autonomous agents and human-machine systems.

Modeling the apparent or latent characteristic appearance of different materials is essential to robustly recognize them in images. 
Early studies of material appearance modeling largely concentrated on comprehensive lab-based measurements using dome systems, robots, or gonioreflectometers collecting measurements that are dense in angular space (such as BRDF, BTF)\cite{Dana2016capturing}. These reflectance-based studies have the advantage of capturing intrinsic invariant properties of the surface, which enables fine-grained material recognition \cite{Zhang_CVPR15,Liu_PAMI14,Salamati_CIC09,Wang_CVPR09}. The inflexibility of lab-based image capture, however, prevents widespread use in real world scenes, especially in the important class of outdoor scenes. 
A fundamentally different approach to reflectance modeling is image-based  modeling where surfaces are captured with a single-view image in-scene or ``in-the-wild.'' 
Recent studies of image-based material recognition use single-view internet-mined images to train classifiers \cite{bell15minc,Hu2011,cimpoi2015deep,Liu2010} and can be applied to arbitrary images casually taken without the need of multiview reflectance information. 
In these methods, however, recognition is typically based more on context than intrinsic material appearance properties except for a few purely local methods \cite{Schwartz_CPCV13,Schwartz_CVPR15}.

Between comprehensive in-lab imaging and internet-mined images, we take an advantageous middle-ground. We capture in-scene appearance but use controlled  viewpoint angles. These measurements  provide a sampling of the full reflectance function. This leads to a very basic question: how do multiple viewing angles help in material recognition?  
Prior work used differential camera motion or object motion for shape reconstruction \cite{chandraker2016information,chandraker2013differential,wang2016svbrdf}, here we consider a novel question:  Do small changes in viewing angles, {\it differential changes}, result in significant increases in recognition performance?
Prior work has shown the power of angular filtering to complement spatial filtering in material recognition. These methods, however, rely on a mirror-based camera to capture a slice of the BRDF \cite{Zhang15} or a lightfield camera to achieve multiple differential viewpoint variations \cite{wang20164d} which limits their application due to the need for specialized imaging equipment.
We instead propose to capture surfaces with differential changes in viewing angles with an ordinary camera and compute discrete approximations of {\it angular gradients} from them. 
We present an approach called {\it angular differential imaging} that augments image 
capture for a particular viewing angle $v$ a differential viewpoint $v+\delta$. Contrast this method with 
lab-based reflectance measurements that often quantize the angular space measuring with domes or positioning devices with large angular spacing such as $22.5^\circ$. These coarse-quantized measurements have limited use in approximating angular gradients. 
Angular differential imaging can be implemented with a small-baseline stereo camera or a moving camera (e.g. handheld).
We demonstrate that differential angular imaging provides key information about material reflectance properties while maintaining the flexibility of convenient in-scene appearance capture.

\begin{table*}[t]
\centering

\begin{tabular}{|l|P{1cm}|P{1cm}|P{.9cm}|P{1.8cm}|P{1.4cm}|P{1.9cm}|P{1.6cm}|l|}
\hline 

Datasets & samples & classes & views & illumination & in scene &scene image& camera parameters& year\\

\hline 
CUReT\cite{dana1999reflectance}&61&61&\multicolumn{2}{P{2.7cm}|}{205}&N&N&N&1999\\

\hline 
KTH-TIPS\cite{hayman2004significance}&11&11&27&3&N&N&N&2004\\

\hline 
UBO2014\cite{weinmann2014material}&84&7&151&151&N&N&N&2014\\

\hline 
Reflectance disk\cite{Zhang15}&190&19&3&3&N&N&Y&2015\\

\hline 
4D Light-field\cite{wang20164d}&1200&12&1&1&Y&N&N&2016\\
\hline 
NISAR\cite{Choe16cvpr}&100&100&9&12&N&N&N&2016\\
\hline 

\textbf{GTOS(ours)}&\textbf{606}&\textbf{40}&\textbf{19}&\textbf{4}&\textbf{Y}&\textbf{Y}&\textbf{Y}&\textbf{2016} \\
         
\hline 
\end{tabular}
\caption{Comparison between GTOS dataset and some publicly available BRDF material datasets. Note that the 4D Light-field dataset\cite{wang20164d} is captured by the Lytro Illum light field camera.}
\label{table:dataset_comparison}
\vspace{-0.15in}
\end{table*}
To capture material appearance in a manner that preserves the convenience of image-based methods and the important angular information of reflectance-based methods, we assemble a comprehensive, first-of-its-kind, {\it outdoor} material database that includes multiple viewpoints and multiple illumination directions (partial BRDF sampling), multiple weather conditions, a large set of surface material classes surpassing existing comparable datasets, multiple physical instances per surface class (to capture intra-class variability) and differential viewpoints to support the framework of differential angular imaging.  
We concentrate on outdoor scenes because of the limited availability of reflectance databases for outdoor surfaces. We also concentrate on materials from ground terrain in outdoor scenes (GTOS) for applicability in numerous application such as automated driving, robot navigation, photometric stereo and shape reconstruction. The 40 surface classes include ground terrain such as grass, gravel, asphalt, concrete, black ice, snow, moss, mud and sand (see Figure~\ref{fig:compare_img_patch}).

We build a recognition algorithm that leverages the strength of deep learning and differential angular imaging.
The resulting method takes two image streams as input, the original image and a differential image as illustrated in Figure~\ref{fig:figure1}.
We optimize the two-stream configuration for material recognition performance and call the resulting network DAIN--differential angular imaging network.

We make three significant contributions in this paper: 1) Introduction of differential angular imaging as a middle-ground between reflectance-based and image-based material recognition; 2) Collection of the GTOS database made publicly available with over 30000 in-scene outdoor images capturing angular reflectance samples with scene context over a large set of material classes; 3) The development of DAIN, a material recognition network with state-of-the-art performance in comprehensive comparative validation.


\section{Related Work}
Texture recognition, the classification of 3D texture images and bidirectional texture functions, traditionally relied on hand-designed 3D image features and multiple views \cite{leung2001representing,cula2001recognition}. More recently, features learned with deep neural networks have outperformed these methods for texture recognition. Cimpoi \etal. \cite{cimpoi2015deep} achieves state-of-art results on FMD \cite{sharan2009material} and KTH-TIPS2 \cite{hayman2004significance} using a Fisher vector representation computed on image features extracted with a CNN. 

The success of deep learning methods in object recognition has also translated to the problem of material recognition, the classification and segmentation of material categories in arbitrary images. Bell \etal., achieve per-pixel material category labeling by retraining the then state-of-the-art object recognition network \cite{simonyan2014two} on a large dataset of material appearance \cite{bell15minc}. This method relies on large image patches that include object and scene context to recognize materials. In contrast, Schwartz and Nishino \cite{Schwartz_CPCV13,Schwartz_CVPR15} learn material appearance models from small image patches extracted inside object boundaries to decouple contextual information from material appearance. To achieve accurate local material recognition, they introduced intermediate material appearance representations based on their intrinsic properties (e.g., ``smooth'' and ``metallic'').

\begin{figure*}
\centering
\subfloat[material classes]
{
  \includegraphics[width=.3\linewidth]{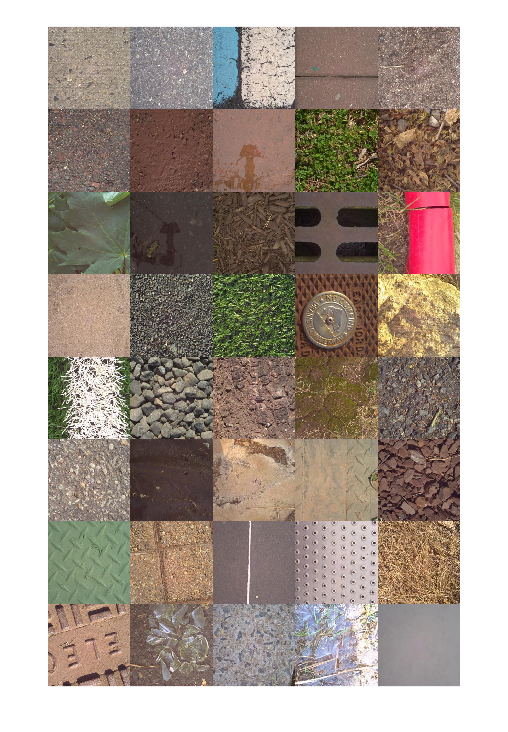}
  
}
\subfloat[one sample at multiple viewing directions]
{
  \includegraphics[width=.65\linewidth]
  {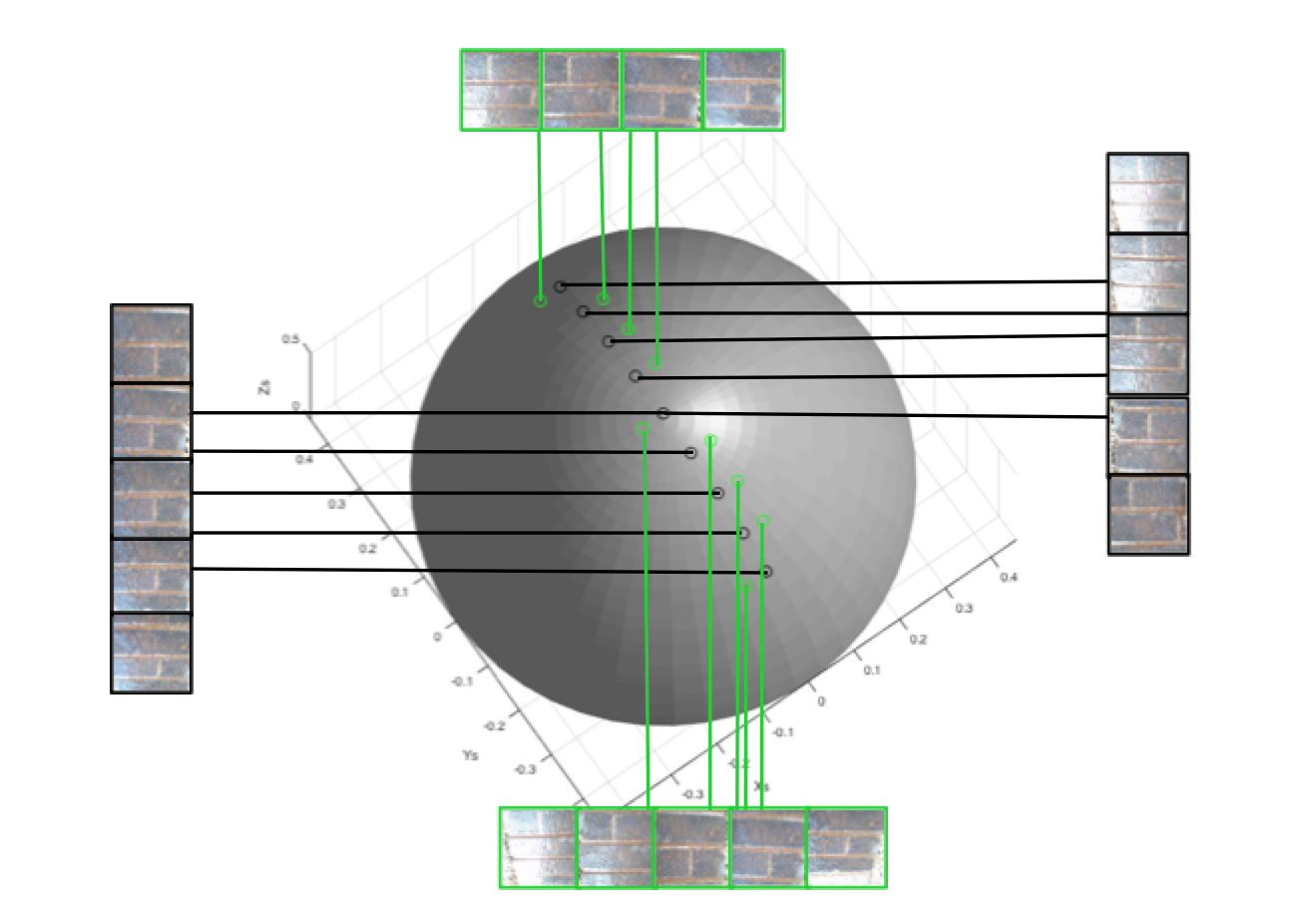}
}
\caption{(a) The 40 material categories in the GTOS dataset introduced in this paper. (Right) The material surface observation points. Nine viewpoint angles separated along an arc spanning $80^\circ$ are measured. For each viewpoint, a differential view is captured $\pm5^\circ$ in azimuth from the original orientation (the sign is chosen based on robotic arm kinematics.)}
\label{fig:database}
\vspace{-0.15in}
\end{figure*}

In addition to the apparent appearance, materials can be discerned by their radiometric properties, namely the bidirectional reflectance distribution function (BRDF) \cite{NicoBRDF} and the bidirectional texture function (BTF) \cite{dana1999reflectance}, which essentially encode the spatial and angular appearance variations of surfaces. Materials often exhibit unique characteristics in their reflectance offering detailed cues to recognize the difference of subtle variations in them (e.g., different types of metal \cite{Liu_PAMI14} and paint \cite{Wang_CVPR09}). Reflectance measurements, however, necessitate elaborate image capture systems, such as a gonioreflectometer \cite{NicoBRDF,Ward92}, robotic arm \cite{lightfield}, or a dome with cameras and light sources \cite{Debevec00,Liu_PAMI14,Wang_CVPR09}.
Recently, Zhang \etal introduced the use of a one-shot reflectance field capture for material recognition \cite{Zhang15}. They adapt the parabolic mirror-based camera developed by Dana and Wang \cite{Dana04} to capture the reflected radiance for a given light source direction in a single shot, which they refer to as a reflectance disk.
More recently, Zhang \etal showed that the reflectance disks contain sufficient information to accurately predict the kinetic friction coeffcient of surfaces \cite{Zhang_ECCV16}. These results demonstrate that the angular appearance variation of materials and their gradients encode rich cues for their recognition. 
Similarly, Wang \etal. \cite{wang20164d} uses a light field camera and combines angular and spatial filtering for material recognition. 
In strong alignment with these recent advances in material recognition, we build a framework of spatial and angular appearance filtering. In sharp contrast to past methods, however, we use image information from standard cameras instead of a multilens array as in Lytro. We explore the difference of using a large viewing angle range (with samples coarsely quantized in angle space) by using differential changes in angles which can easily be captured by a two-camera system or small motions of a single ordinary camera. 

Deep learning has achieved major success in  object classification \cite{krizhevsky2012imagenet,Chatfield14,he2015deep}, segmentation \cite{girshick2014rich,karpathy2015deep,ren2015faster}, and material recognition \cite{cimpoi2015deep,Zhang_ECCV16,lin2015bilinear,zhang2016deep}.
In our goal of combining spatial and angular image information to account for texture and reflectance, we 
 are particularly motivated by the
 two-stream fusion framework \cite{feichtenhofer2016convolutional,simonyan2014two} which achieves state-of-art results in UCF101\cite{soomro2012ucf101} action recognition dataset.
%
%
%
\vspace{-0.1in}
\paragraph{\bf Datasets:}
Datasets to measure reflectance of real world surfaces have a long history of lab-based measurements including: CUReT database\cite{dana1999reflectance}, KTH-TIPS database by Hayman \etal. \cite{hayman2004significance}, MERL Reflectance Database \cite{Matusik03}, UBO2014 BTF Database \cite{weinmann2014material}, UTIA BRDF Database \cite{Filip14}, Drexel Texture Database \cite{Oxholm_ECCV12-2} and IC-CERTH Fabric Database \cite{Kampouris16}.
In many of these datasets, dense reflectance angles are captured with special image capture equipment. 
Some of these datasets have limited instances/samples per surface category (different physical samples representing the same class for intraclass variability) or have few surface categories, and all are obtained from indoor measurements where the sample is removed from the scene. 
More recent datasets capture materials and texture in-scene, (a.k.a. in-situ, or in-the-wild). A motivation of moving to in-scene capture is to build algorithms and methods that are more relevant to real-world applications. 
These recent databases are from internet-mined databases and contain a single view of the scene under a single illumination direction. Examples include the the Flickr Materials Database by Sharan \etal. \cite{sharan2009material} and the Material in Context Database by Bell \etal. \cite{bell15minc}.
%
Recently, DeGol \etal released GeoMat Database\cite{degol2016geometry} with 19  material categories  from outdoor sites and 
each category has between 3 and 26 physical surface instances, with 8 to 12 viewpoints per surface. The viewpoints in this dataset are  irregularly sampled in angle space.

\section{Differential Angular Imaging}
We present a new measurement method called differential angular imaging where a surface is imaged from a particular viewing angle $v$ and then from an additional viewpoint $v+\delta$. The motivation for this differential change in viewpoint is improved computation of the angular gradient of intensity $\partial{I_v}/\partial{v}$. Intensity gradients are the basic building block of image features and it is well known that discrete approximations to derivatives have limitations. In particular, spatial gradients of intensities for an image $I$ are approximated by $I(x+\Delta)-I(x)$ and this approximation is most reasonable at low spatial frequencies and when $\Delta$ is small. 
For angular gradients of reflectance, the discrete approximation to the derivative is a subtraction with respect to the viewing angle. Angular gradients are approximated by $I(v+\delta)-I(v)$ and this approximation  requires a small $\delta$. Consequently, differential angular imaging provides more accurate angular gradients.  

The differential images as shown in Figures~\ref{fig:figure1} and \ref{fig:compare_img_patch} have several characteristics. First, the differential image reveals the gradients in BRDF/BTF at the particular viewpoint. Second, relief texture is also observable in the differential image due to non-planar surface structure. Finally, the differential images are sparse. This sparsity can provide a computational advantage within the network. (Note that $I(v+\delta)$ and $I(v)$ are aligned with a global affine transformation before subtraction.)   

\vspace{-0.08in}
\section{GTOS Dataset}
\begin{figure}
\centering
\includegraphics[width= \linewidth]{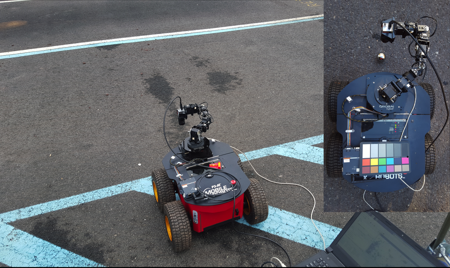}
\caption{The measurement equipment for the GTOS database: Mobile Robots P3-AT robot, Cyton gamma 300 robot arm, Basler aca2040-90uc camera with Edmund Optics 25mm/F1.8 lens, DGK 18\% white balance and color reference card, and Macmaster-Carr 440C Stainless Steel Sphere.}
\label{fig:equipment}
\vspace{-0.2in}
\end{figure}

\paragraph{Ground Terrain in Outdoor Scenes Dataset}
We collect the GTOS database, a first-of-its-kind in-scene material reflectance database, to investigate the use of spatial and angular reflectance information of outdoor ground terrain for material recognition. 
We capture reflectance systematically by imaging a set of viewing angles comprising a partial BRDF with a mobile exploration robot. Differential angular images are obtained by also measuring each of $N_v=9$ base angles $v=(\theta_v,\phi_v)$, $\theta_v \in [-40^\circ,-30^\circ,\dots,40]$, and a differential angle variation of $\delta = (0,5^\circ)$ resulting in 18 viewing directions per sample as shown in Figure~\ref{fig:database} (b).
Example surface classes are depicted in Figure~\ref{fig:database} (a).
The class names are (in order of top-left to bottom-right): cement, asphalt, painted asphalt , brick, soil, muddy stone, mud, mud-puddle, grass, dry leaves, leaves, asphalt-puddle, mulch, metal grating, plastic, sand, stone, artificial turf, aluminum,  limestone, painted turf, pebbles, roots, moss, loose asphalt-stone, asphalt-stone, cloth, paper, plastic cover, shale, painted cover, stone-brick, sandpaper, steel, dry grass, rusty cover, glass, stone-cement, icy mud, and snow. The $N_c=40$ surface classes mostly have between 4 and 14 instances (samples of intra-class variability) and each instance is imaged not only under $N_v$ viewing directions but also under multiple natural light illumination conditions. As illustrated in Figure~\ref{fig:figure1}, sample appearance depends on the weather condition and the time of day.  
To capture this variation, we image the same region with $N_i=4$ different weather conditions (cloudy dry, cloudy wet, sunny morning, and sunny afternoon).
We capture the samples with 3 different exposure times to enable high dynamic range imaging. Additionally, we image a mirrored sphere
to capture the environment lighting of the natural sky. 
In addition to surface images, we capture a scene image to show the global context. 
The robot measurement device is depicted in Figure ~\ref{fig:equipment}. Although, the database measurements were obtained with robotic positioning for precise angular measurements, our recognition results are based on subsets of these measurements so that an articulated arm would not be required for an in-field system.
The total number of surface images in the database is 34,243. 
As shown in Table~\ref{table:dataset_comparison}, this is the most extensive outdoor in-scene multiview material database to date.




\section{DAIN for Material Recognition}
\paragraph{Differential Angular Imaging Network (DAIN)}
\begin{figure*}
\centering
\subfloat[Final layer (prediction) combination method]
{
  \includegraphics[width=.4\linewidth]{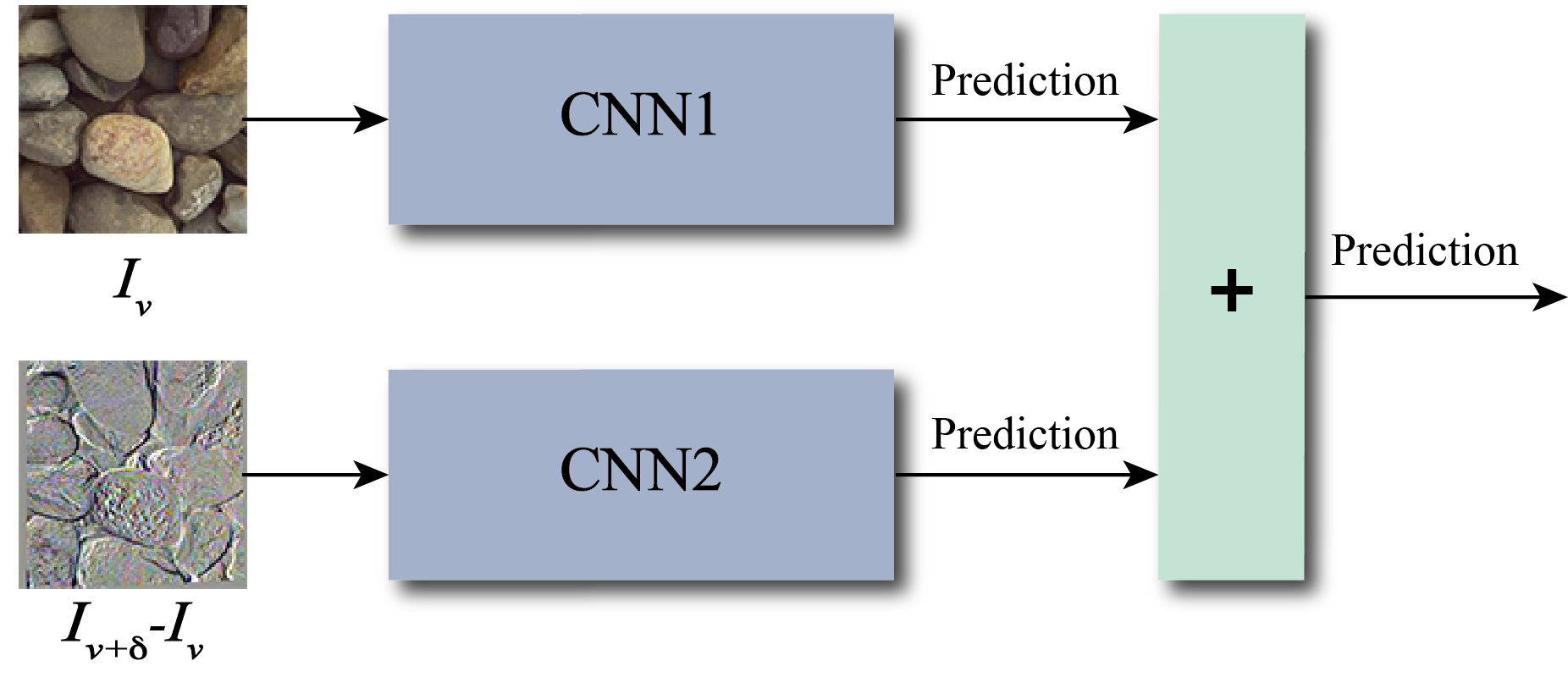}
  
}
\subfloat[Intermediate layer (feature maps) combination method]
{
  \includegraphics[width=.42\linewidth]
  {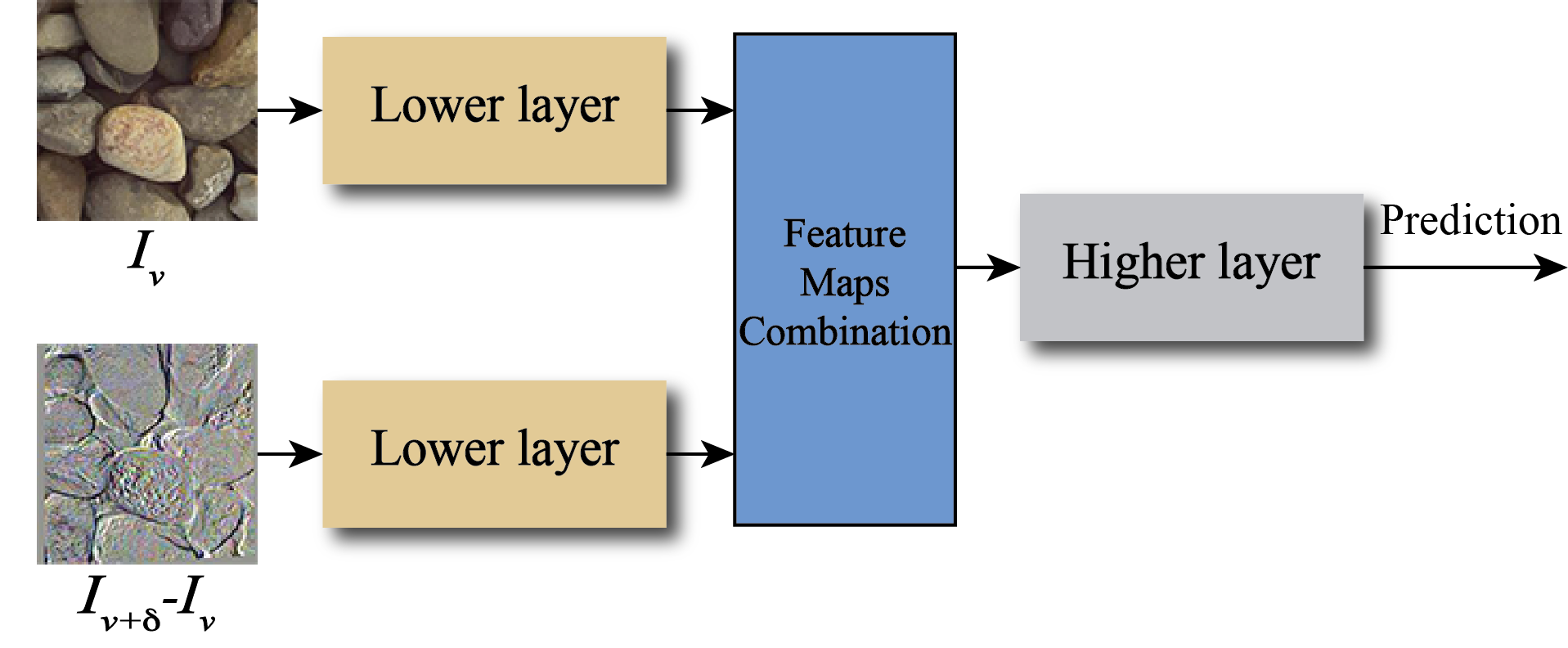}
}

\subfloat[DAIN (differential angular image network)]
{
  \includegraphics[width=.6\linewidth]
  {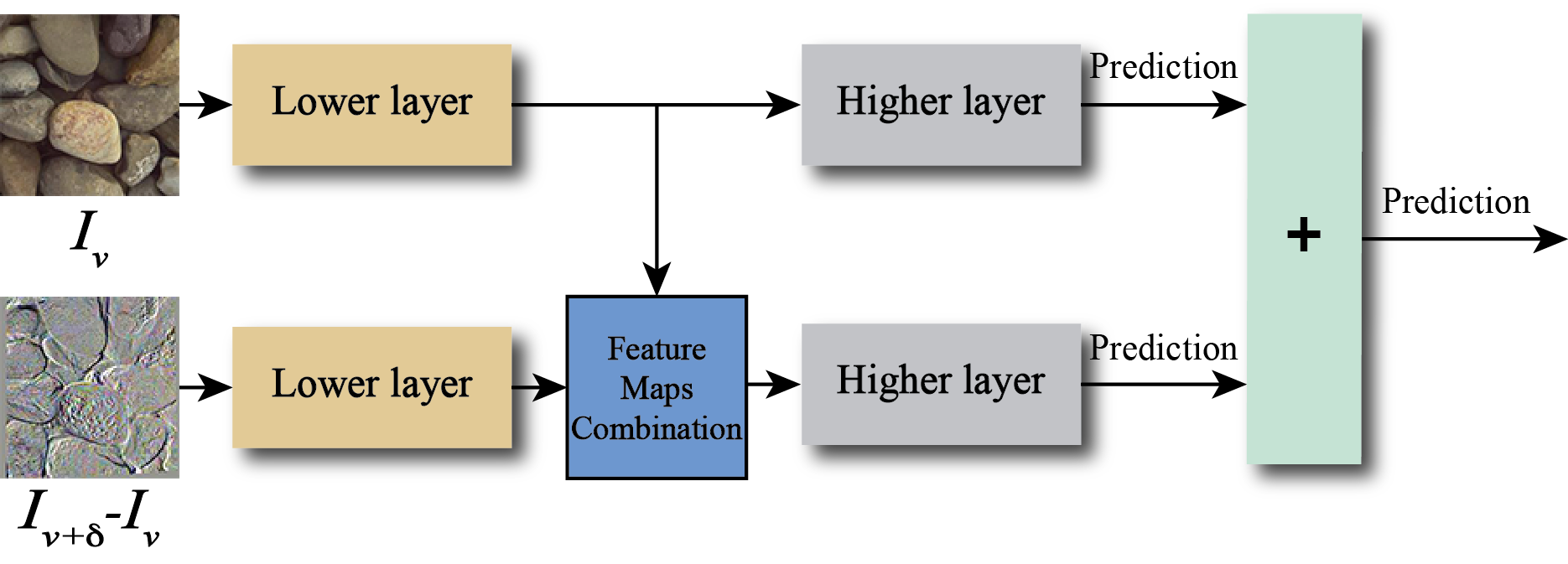}
}
\caption{Methods to combine two image streams, the original image $I_v$ and the differential image $I_\delta= I_{v+\delta}-I_v$. The best performing configuration is the architecture in (c), which we refer to as differential angular imaging network (DAIN).}
\label{fig:fuse_cmp}
\vspace{-0.15in}
\end{figure*}
Consider the problem of in-scene material recognition with images from multiple viewing directions (multiview). We develop a two-stream convolutional neural network to fully leverage differential angular imaging for material recognition. 
The differential image $I_\delta$ sparsely encodes reflectance angular gradients as well as surface relief texture. The spatial variation of image intensity remains an important recognition cue and so our method integrates these two streams of information. 
A CNN is used on both streams of the network and then combined for the final prediction result. The combination method and the layer at which the combination takes place leads to variations of the architecture.

We employ the ImageNet \cite{deng2009imagenet} pre-trained VGG-M model \cite{Chatfield14} as the prediction unit (labeled CNN in Figure~\ref{fig:fuse_cmp}). The first input branch is the image  $I_v$ at a specific viewing direction $v$. The second input branch is the differential image $I_\delta$. 
The first method of combination  shown in Figure~\ref{fig:fuse_cmp} (a) is a simple averaging of the output prediction vectors obtained by the two branches. The second method  combines the two branches at the intermediate layers of the CNN, i.e. the feature maps output at layer $M$ are combined and passed forward to the higher layers of the CNN, as shown Figure~\ref{fig:fuse_cmp} (b).
We empirically find that combining feature maps generated by Conv5 layer after ReLU performs best.
A third method (see Figure~\ref{fig:fuse_cmp} (c)) is a hybrid of the  two architectures that preserves the original CNN path for the original image $I_v$ by combining the layer $M$ feature maps for both streams {\it and} by combining the prediction outputs for both streams as shown in Figure~\ref{fig:fuse_cmp} (c). This approach is the best performing architecture of the three methods and we call it the differential angular imaging network (DAIN). 

For combining feature maps at layer $M$,
consider features maps ${x}_{a}$ and $x_b$ from the two branches 
that have width $W$, height $H$, and feature channel depth $D$. 
The output feature map $y$ will be the same dimensions  ${W \times H \times D}$.
We can combine feature maps by: (1) {\it Sum:} pointwise sum of ${x}_{a}$ and $x_b$, and (2) {\it Max:} pointwise maximum of ${x}_{a}$ and $x_b$.
In Section~\ref{sec:experiments} we evaluate the performance of these methods of combining lower layer feature maps.

\subsection{Multiple Views}
Our GTOS database has multiple viewing directions on an arc (a partial BRDF sampling) as well as differential images for each viewing direction. 
We evaluate  our recognition network in two modes:
(1) {\bf Single view DAIN}, with inputs from $I_v$ and $I_\delta$, with $v$ representing a single viewing angle; (2) {\bf Multi view DAIN}, with inputs $I_v$ and $I_\delta$, with $v \in [v1,v2,...,vN]$. For our GTOS databse,  $v1,v2,...,vN$ are viewing angles separated by $10^\circ$ representing a $N \times 10^\circ$ range of viewing angles. We empirically determine that $N=4$ viewpoints are sufficient for recognition. 
For a baseline comparison we also consider non-differential versions: {\bf Single View} with only $I_v$ for a single viewing direction and {\bf Multi View} with inputs $I_v$,   $v \in [v1,v2,...,vN]$.

\begin{figure*}
\centering
\includegraphics[width= .95\textwidth]{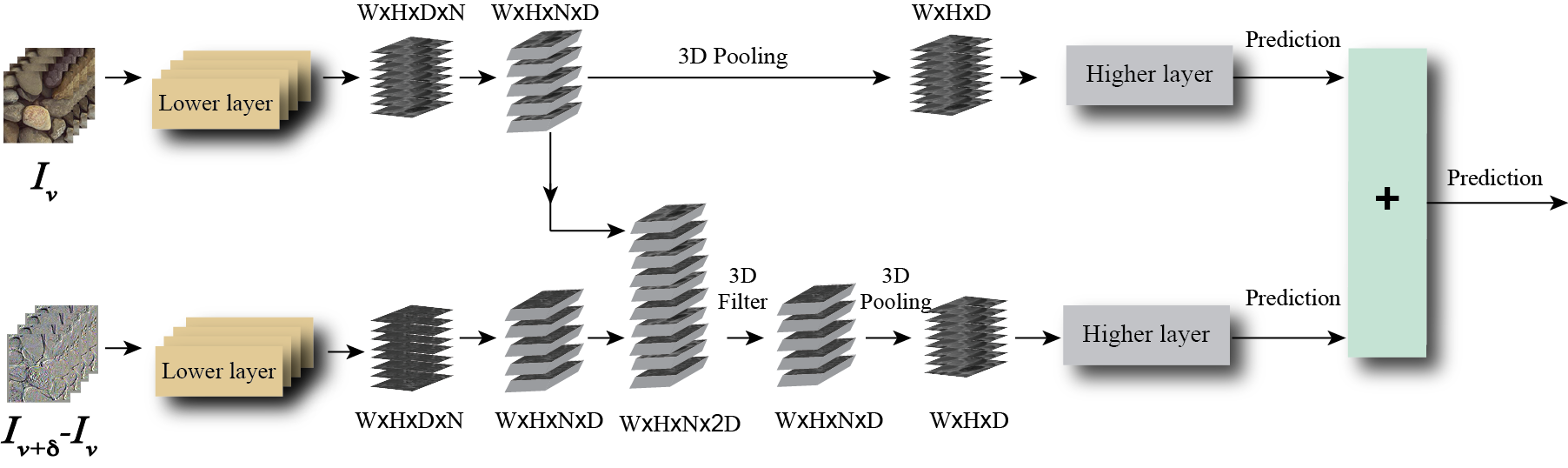}
\caption{Multiview DAIN. The 3D filter + pooling method to combine two streams (original and differential image) from multiple viewing angles.   $W$, $H$, and $D$ are the width, height, and depth of corresponding feature maps, $N$ is the number of view points.}
\label{fig:3D_filter}
\vspace{-0.2in}
\end{figure*}

To incorporate multi view information in DAIN we use three methods:
(1) voting (use the predictions from each view to vote), 
(2) pooling (pointwise maximum of the combined feature maps across viewpoints),
(3) 3D filter + pooling (follow \cite{tran2015learning} to use a $3\times 3 \times 3$ learned filter bank to convolve the multi view feature maps). See Figure~\ref{fig:3D_filter}. 
After 3D filtering, pooling is used (pointwise maximum across viewpoints). 
The computational expense of this third method due to learning the filter weights is significantly higher. 
\section{Experiments}
\label{sec:experiments}
In this section, we evaluate the DAIN framework for 
 material recognition and compare the results on GTOS with several state-of-the-art algorithms. 
The first evaluation determines which structure of the two stream networks from Figure~\ref{fig:fuse_cmp} works best on the GTOS dataset, leading to the choice in (c) as the DAIN architecture. The second evaluation considers recognition performance with different variations of DAIN recognition. 
The third experimental evaluation compares three other state-of-the-art approaches on our GTOS-dataset, concluding that multiview DAIN works best. 
Finally, we apply DAIN to a lightfield  dataset to show performance in another multiview material dataset. 
 \vspace{-0.2in}
 \paragraph{Training procedure}
We design 5 training and testing splits by assigning about 70\% of ground terrain surfaces of each class to training and the rest 30\% to testing.
Note that, to ensure that there is no overlap between training and testing sets, if one sample is in the training set, all views and illumination conditions for that sample is in the training set. 

 Each input image from our GTOS database is resized into 240 $\times$ 240.
Before training a two branch network, we first  fine-tune the VGG-M model separately with original and differential images with batch size 196, dropout rate 0.5, momentum 0.9. 
We employ the augmentation method that horizontally and vertically stretch training images within $\pm10\%$, with an optional 50\% horizontal mirror flips.
The images are randomly cropped into 224 $\times$ 224 material patches.
All images are pre-processed by subtracting a per color channel mean and normalizing for unit variance.
The learning rate for the last fully connected layer is set to 10 times of other layers.
We first fine-tune only the last fully connected layer with learning rate $5 \times {10}^{-2}$ for 5 epochs;
then, fine-tune all the fully connected layers with learning rate ${10}^{-2}$ for 5 epochs.
Finally we fine-tune all the layers with leaning rate starting at ${10}^{-3}$, and decrease by a factor of 0.1 when the training accuracy saturates.
Since the snow class only has 2 samples, we omit them from experiments.

For the two branch network, we employ the fine-tuned two-branch VGG-M model with batch size 64 and learning rate starting from ${10}^{-3}$ which is reduced by a factor of 0.1 when the training accuracy saturates.
We augment training data with randomly stretch training images by $\pm25\%$ horizontally and vertically, and also horizontal mirror flips.
The images are randomly cropped to 224 $\times$ 224 material patches.
We first backpropagate only to feature maps combination layer for 3 epochs, then fine tunes all layers.
We employ the same augmentation method for the multiview images of each material surface.
We randomly select the first viewpoint image, then subsequent $N=4$ view point images are selected for experiments.
\vspace{-0.1in}
\paragraph{Evaluation for DAIN Architecture}
Table~\ref{table:sum_comparison} shows the mean classification accuracy of the different three branch combination methods depicted in Figure~\ref{fig:fuse_cmp}. 
Inputs are single view images (${I}_{v}$) and single view differential images (${I}_{\delta}$).
Combining the two streams at the final prediction layer  (77\% accuracy) is compared with the intermediate layer combination  (74.8\%) or the hybrid approach in Figure~\ref{fig:fuse_cmp} (c) (79.4\%) which we choose as the differential angular imaging network. 
The combination method used is Sum and the feature maps are obtained from  Conv5 layers after ReLU.


\vspace{-0.1in}
\paragraph{DAIN Recognition Performance}
We evaluate DAIN recognition performance for single view input (and differential image) and for multiview input from the GTOS database. Additionally, we compare the results to recognition using a standard CNN without a differential image stream. For all multiview experimental results we choose the number of viewpoints $N=4$, separated by $10^\circ$ with the starting viewpoint chosen at random (and the corresponding differential input). 
Table~\ref{table:method_comparison} shows the resulting recognition rates (with standard deviation over 5 splits shown as a subscript).  
The first three rows shows the accuracy {\it without} differential angular imaging, using both single view and multiview input. Notice the recognition performance for these non-DAIN results are generally lower than the DAIN recognition rates in the rest of the table. 
The middle three rows show the recognition results for single view DAIN.
For combining feature maps we evaluate both Sum and Max which have comparable results. 
Notice that single view DAIN achieves better recognition accuracy than multiview CNN with voting (79.4\% vs. 78.1\%). This is an important result indicating the power of using the differential image. Instead of four viewpoints separated by $10^\circ$ a single viewpoint and its differential image achieves a better recognition. These results provide design cues for  building imaging systems tailored to material recognition.
We also evaluate weather using inputs from the two viewpoints directly (i.e. $I_v$ and $I_{v+\delta}$) is comparable to using $I_v$ and the differential image $I_{\delta}$.
Interestingly, the differential image as input has an advantage (79.4\% over 77.5\%).
The last three rows of Table~\ref{table:method_comparison} 
show that recognition performance using multiview DAIN beats the performance of both single view DAIN and CNN methods with no differential image stream.
We evaluate different ways to combine the multiview image set including voting, pooling, and the 3D filter+pooling illustrated in Figure~\ref{fig:3D_filter}.



The CNN module of our DAIN network can be replaced by other state-of-the-art deep learning methods to further improve results. To demonstrate this, we change the CNN module in a single view DAIN (Sum) (with inputs ${I}_{v}$,  ${I}_{\delta}$) to ImageNet pre-trained ResNet-50 model\cite{he2015deep} on split1. Combining feature maps generated from the Res4 layer (the fourth residual unit) after ReLU with training batch size 196, 
recognition rate improves from 77.5\% to 83.0\%.

Table~\ref{table:state_art_comparison} shows the recognition rates for
multiview DAIN that outperforms three other multi-view classification method:
 FV+CNN\cite{cimpoi2014describing}, FV-N+CNN+N\tiny3D \normalsize\cite{degol2016geometry}, and MVCNN\cite{su2015multi}.
The table shows recognition rates for a single split of the GTOS database 
with images resized to 240 $\times$ 240.
All experiments are based on the same pre-trained VGG-M model. 
We use the same fine-tuning and training procedure as in the MVCNN\cite{su2015multi} experiment.
For FV-N+CNN+N\tiny3D \normalsize applied to GTOS, 10 samples (out of 606) failed to get geometry information by the method provided in \cite{degol2016geometry} and we removed these samples from the experiment. 
The patch size in \cite{degol2016geometry} is 100 $\times$ 100, but the accuracy for this patch size for GTOS was only 43\%, so we use 240 $\times$ 240. 
We implement FV-N+CNN+N\tiny3D \normalsize with linear mapping instead of homogeneous kernel map\cite{vedaldi2012efficient} for SVM training to save memory with this larger patch size.
\vspace{-0.1in}
\paragraph{DAIN on 4D Light Field Dataset} 
We tested our multiview DAIN (Sum + pooling) method on a recent 4D light field (Lytro) dataset \cite{wang20164d}. ResNet-50 is used as the CNN module. 
The recognition accuracy with full images on 5 splits is 83.0\tiny $\pm2.1$ \normalsize . Note that a subset  of  the lightfield data is used to mimic the differential imaging process, so these results should not be interpreted as a comparison of our algorithm to \cite{wang20164d}. 
 

The Lytro dataset has $N=49$ views, from the 7 $\times$ 7 lenslet array, where each lenslet corresponds to a different viewing direction. 
Using $(i,j)$ as an index into this array, we employ the viewpoints
indexed by $(4,1), (4,3), (4,5), (4,7)$ as the 4 views in multiview DAIN. We use the viewpoint indexed by $(3,1), (5,3), (3,5), (5,7)$ as the corresponding differential views. This is an approximation of multiview DAIN; the lightfield dataset does not capture the range of viewing angles to exactly emulate multiple viewpoints and small angle variations of these viewpoints.
Instead of using all $N=49$ viewpoints as in \cite{wang20164d}, we generate comparable recognition accuracy by only 8 viewpoints.

\begin{table}
\centering
\begin{tabular}{|l|M{1.5cm}|M{2cm}|M{1.5cm}|l|}
\hline 

 Method & Final Layer Combination  & Intermediate Layer Combination  & DAIN \\

\hline 
Accuracy&77.0\tiny $\pm2.5$&74.8\tiny $\pm3.4$&79.4\tiny $\pm3.4$ \\
          
\hline 
\end{tabular}
\caption{Comparison of accuracy from different two stream methods as shown in Figure~\ref{fig:fuse_cmp}. The feature-map combination method for (b) and (c) is Sum at Conv5 layers after ReLU. The reported result is the mean accuracy and the subscript shows the standard deviation over 5 splits of the data. Notice that the architecture in (c) gives the best performance and is chosen for the differential angular imaging network (DAIN).}
\label{table:sum_comparison}
\vspace{-0.15in}
\end{table}



\begin{table}
\centering

\resizebox{\linewidth}{!}{%
\begin{tabular}{|l|M{1cm}|M{1.4cm}|l|}
\hline 

Method & First input & Second input & Accuracy\\

\hline 
single view CNN &${I}_{v}$&-&74.3\tiny$\pm2.8$\\
 
multiview CNN, voting&${I}_{v}$&-&78.1\tiny$\pm2.4$\\

multiview CNN,3D filter&${I}_{v}$&-&74.8\tiny$\pm3.2$\\

\hline 

single view DAIN (Sum)&${I}_{v}$&${I}_{v+\delta}$&77.5\tiny $\pm2.7$\\
 
single view DAIN (Sum)&${I}_{v}$&${I}_{\delta}$&79.4\tiny $\pm3.4$\\

single view DAIN (Max)&${I}_{v}$&${I}_{\delta}$&79.0\tiny $\pm1.8$\\


\hline 

multiview DAIN (Sum/voting)&${I}_{v}$&${I}_{\delta}$&80.0\tiny $\pm2.1$\\

multiview DAIN (Sum/pooling)&${I}_{v}$&${I}_{\delta}$&81.2\tiny $\pm1.7$\\


multiview DAIN (3D filter/pooling)&${I}_{v}$&${I}_{\delta}$&81.1\tiny $\pm1.5$\\
         
\hline 

\end{tabular}
}

\caption{Results comparing performance of standard CNN recognition without angular differential imaging (first three rows) to our single-view DAIN (middle three rows) and our multi-view DAIN (bottom three rows). 
 ${I}_{v}$ denotes the image from viewpoint $v$, $I_{v+\delta}$ is the image obtained from viewpoint $v+\delta$, and $I_\delta=I_v - {I}_{v+\delta}$ is the differential image. The differential angular imaging network (DAIN) has superior performance over CNN even when comparing single view DAIN to multiview CNN.
Multiview DAIN provides the best recognition rates.}
\label{table:method_comparison}
\vspace{-0.1in}
\end{table}

\begin{table}
\centering

\begin{tabular}{|l|l|}
\hline 

 Architecture & Accuracy\\

\hline 
FV+CNN\cite{cimpoi2014describing}&75.4\%\\
 
FV-N+CNN+N\tiny3D \normalsize\cite{degol2016geometry}&58.3\%\\

MVCNN\cite{su2015multi}&78.1\%\\

\textbf{multiview DAIN (3D filter),  pooling}&\textbf{81.4\%}\\
         
\hline 
\end{tabular}
\caption{Comparison with the state of art algorithms on GTOS dataset. Notice that our method, multiview DAIN, achieves the best recognition accuracy. 
}
\label{table:state_art_comparison}
\vspace{-0.2in}
\end{table}

\vspace{-0.05in}
\section{Conclusion}
In summary, there are three main contributions of this work: 1) Differential Angular Imaging for a sparse spatial distribution of angular gradients that provides key cues for material recognition; 2) The GTOS Dataset with ground terrain imaged by systematic in-scene measurement of partial reflectance instead of in-lab reflectance measurements. The database contains 34,243 images with 40 surface classes, 18 viewing directions, 4 illumination conditions, 3 exposure settings per sample and several instances/samples per class.
3) We develop and evaluate an architecture for using differential angular imaging, showing superior results for differential inputs as compared to original images. 
Our work in measuring and modeling outdoor surfaces has important implications for applications such as robot navigation (determining control parameters based on current ground terrain) and automatic driving (determining road conditions by partial real time reflectance measurements). We believe our database and methods will provide a sound foundation for in-depth studies on material recognition in the wild.

\vspace{-0.1in}
\section*{Acknowledgment}
This work was supported by National Science Foundation 
award IIS-1421134. A GPU used for this research was donated by the NVIDIA Corporation.
Thanks to Di Zhu, Hansi Liu, Lingyi Xu, and Yueyang Chen for help with data collection.

{\small
\bibliographystyle{ieee}
\bibliography{egbib}
}

\end{document}